\documentclass[10pt,twocolumn,letterpaper]{article}

\usepackage{iccv}              

\usepackage[accsupp]{axessibility}

\usepackage{algorithm}
\usepackage{algpseudocode}
\usepackage{amsmath}
\usepackage{caption}
\usepackage{placeins}

\usepackage{algorithm}
\usepackage{algpseudocode}
\usepackage{amsmath}
\usepackage{multirow}
\usepackage{wrapfig}
\usepackage{graphicx}
\usepackage[most]{tcolorbox}
\usepackage{array}
\usepackage{booktabs}         
\usepackage{tabularx} 
\usepackage[utf8]{inputenc}
\usepackage{textgreek}
\usepackage{makecell}   
\usepackage{adjustbox}  
\usepackage{makecell}   
\usepackage{array}      
\usepackage{adjustbox}  
\usepackage{booktabs}   
\usepackage{graphicx}
\usepackage{multirow, array}
\usepackage{float} 
\usepackage[accsupp]{axessibility}  

\definecolor{iccvblue}{rgb}{0.21,0.49,0.74}
\usepackage[pagebackref,breaklinks,colorlinks,allcolors=iccvblue]{hyperref}

\def\paperID{*****}
\def\confName{ICCV}
\def\confYear{2025}

\newif\ifSUPP
\SUPPfalse    

\ifSUPP
  \title{How Well Do Vision--Language Models Understand Cities?\\
  A Comparative Study on Spatial Reasoning from Street-View Images}
  \author{} 
\else
  \title{How Well Do Vision--Language Models Understand Cities?\\
  A Comparative Study on Spatial Reasoning from Street-View Images}
  \author{
    Juneyoung Ro \quad Namwoo Kim \quad Yoonjin Yoon$^{*}$ \\
    Korea Advanced Institute of Science and Technology \\
    {\tt\small \{juneyoung, namwoo, yoonjin\}@spacetime.kaist.ac.kr}
  }
\fi

\begin{document}
\maketitle

\ifSUPP
\else
  \begingroup
  \renewcommand{\thefootnote}{\fnsymbol{footnote}}
  \footnotetext[1]{Corresponding author: yoonjin@spacetime.kaist.ac.kr}
  \endgroup

  \begin{abstract}
Effectively understanding urban scenes requires fine-grained spatial reasoning about objects, layouts, and depth cues. However, how well current vision-language models (VLMs), pretrained on general scenes, transfer these abilities to urban domain remains underexplored. To address this gap, we conduct a comparative study of three off-the-shelf VLMs—BLIP-2, InstructBLIP, and LLaVA-1.5—evaluating both zero-shot performance and the effects of fine-tuning with a synthetic VQA dataset specific to urban scenes. We construct such dataset from segmentation, depth, and object detection predictions of street-view images, pairing each question with LLM-generated Chain-of-Thought (CoT) answers for step-by-step reasoning supervision. Results show that while VLMs perform reasonably well in zero-shot settings, fine-tuning with our synthetic CoT-supervised dataset substantially boosts performance, especially for challenging question types such as negation and counterfactuals. This study introduces urban spatial reasoning as a new challenge for VLMs and demonstrates synthetic dataset construction as a practical path for adapting general-purpose models to specialized domains. 
\end{abstract}

  \vspace{-10pt}
\section{Introduction}
\label{sec:intro}

\noindent
Understanding street-level urban scenes at fine-grained spatial scales is crucial for informing how cities are designed and experienced. Visual elements such as greenery, skyline openness, and building density significantly influence urban comfort, walkability, and perceived safety \cite{Naik2014,Bardhan2024,Dubey2016,Duarte2024}. While humans can intuitively grasp these features when viewing street-level imagery, current AI models—particularly those designed for general-purpose visual understanding—often face difficulties in reasoning about spatial relationships and compositional patterns in complex urban environments.

\noindent
\begin{figure}[ht]  
  \centering

  \includegraphics[width=1.0\linewidth]{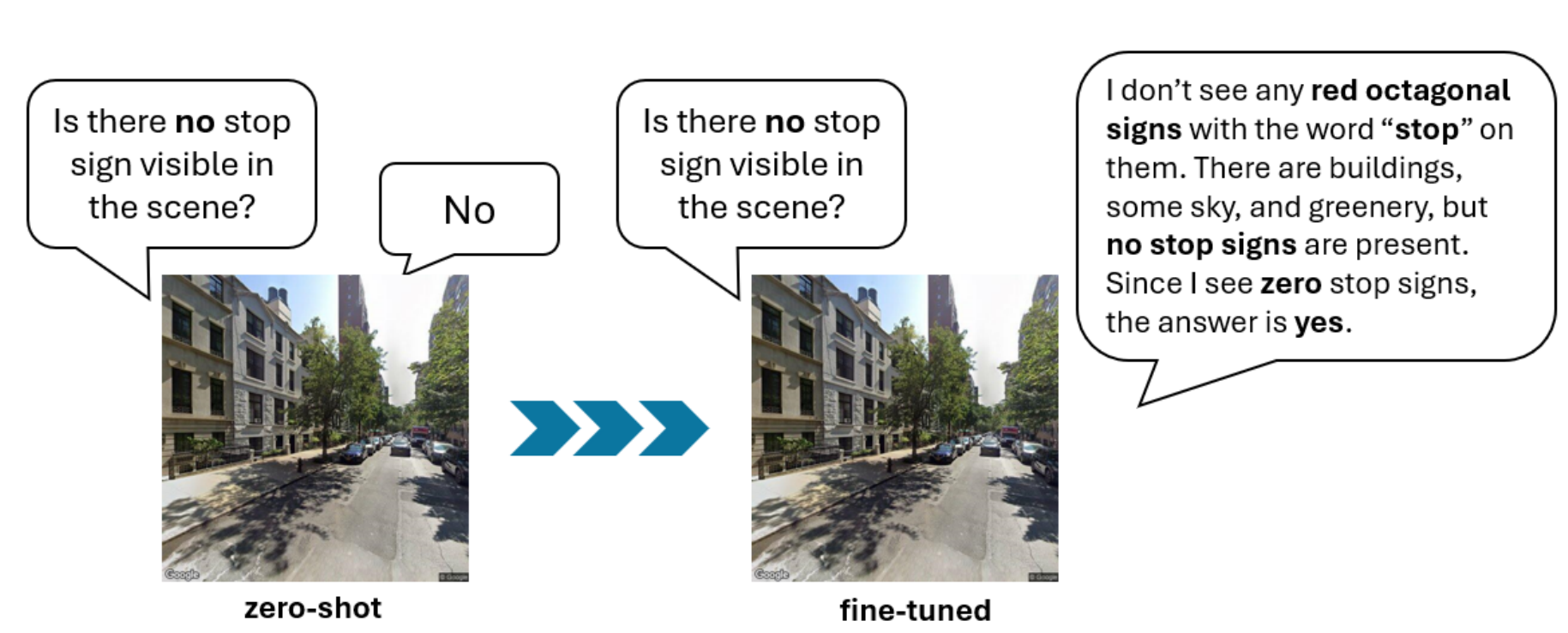}
  \caption{Example illustrating a zero-shot model's failure in handling spatial reasoning under negation, contrasted with a fine-tuned model’s correct, step-by-step spatial reasoning process.}
  \label{fig:teaser-image}
\end{figure}

\noindent 
\textbf{Out-of-Domain Challenge in Urban Scenes.}  
Large-scale models like CLIP \cite{Radford2021} excel on images resembling those in their pretraining corpora, yet frequently underperform on domains such as medical scans, satellite imagery, and “natural adversarial” photos that differ in texture, structure, or composition \cite{Li2025,Hendrycks2021}. A recent survey \cite{Li2025} attributes this limitation to the model's tendency to rely on surface-level keyword associations rather than reasoning about spatial structures and relationships.
Urban street scenes are particularly affected by these issues. While they share recurring elements, such as trees and buildings, fine-grained cues such as how much tall buildings block the sky, the canopy area of trees that creates shaded rest spots, or the overall greenery index that signals biophilic quality all convey crucial spatial information. Accurately interpreting these details is vital for evidence-based planning \cite{Hu2024}, navigation \cite{He2024}, and public-space design \cite{Chen2024b}.

Street-view imagery serves as a primary tool for capturing and analyzing such fine-grained urban features. It has been extensively employed to study human-inhabited urban environments \cite{Nagata2020, Li2015}. Recent research has focused on streetscape perception evaluation \cite{Ogawa2024, Liang2024}, and the integration of reasoning module may further enhance the depth of visual urban analysis. However, existing studies often emphasize perceptual assessment without systematically incorporating reasoning about spatial relationships. To address this gap, our pipeline generates both \textbf{perceptual} and \textbf{compositional} base QA pairs to assess how well state-of-the-art vision–language models interpret urban scenes. We then transform these base pairs into Chain-of-Thought (CoT) variants that verbalize the underlying reasoning process, supporting evaluation of final answers alongside reasoning fidelity. This design allows even simple yes/no questions to probe fine-grained spatial understanding and the ability to articulate metadata-grounded inferences.

\smallskip
In pursuit of this goal, we introduce a synthetic QA generation pipeline built to fine-tune vision–language models (VLMs) for urban street-scene understanding. Leveraging segmentation, depth estimation, and object detection predictions, we construct thousands of visually grounded questions and answers, providing targeted supervision for spatial and compositional understanding. We use this synthetic dataset to evaluate three off-the-shelf VLMs—BLIP-2~\cite{Li2023}, InstructBLIP~\cite{Dai2023}, and LLaVA-1.5~\cite{Liu2024}—covering the spectrum from zero-shot transfer models to fully conversational multimodal assistants. These models were selected for their architectural diversity, distinct training strategies, and full open-source availability, which enables transparent inspection, reproducibility, and community-driven improvement. BLIP-2 serves as a strong general-purpose model with frozen vision encoders, InstructBLIP as an instruction-tuned variant optimized for multimodal following, and LLaVA-1.5 as a large-scale, conversational VLM with end-to-end fine-tuning. By comparing each model in its original form and after fine-tuning, we assess whether modest adaptation can yield reliable, human-aligned street-level reasoning. Below are our key research questions and contributions.

\vspace{0.4em}
\noindent
\textbf{Research Questions}  
\begin{itemize}
    \item \textbf{RQ1} – How well do current vision-language models understand fine-grained spatial relationships in urban street scenes out of the box?
    \item \textbf{RQ2} – How much can targeted fine-tuning on a synthetic but carefully structured domain-specific QA set close this gap?
    \item \textbf{RQ3} – How do model strengths and weaknesses vary across different question types, from perception-based to compositional reasoning tasks?
\end{itemize}

\smallskip
\noindent
\textbf{Our Contributions}
\begin{itemize}
\item We conduct a comparative study of BLIP-2, InstructBLIP, and LLaVA-1.5 on fine-grained spatial reasoning in urban street scenes, enabled by a synthetic VQA dataset we construct from street-view images to support both zero-shot and fine-tuned evaluation.
\item We develop a modular QA generation pipeline that produces 280K questions across perceptual, compositional, and CoT formats, enabling diverse and progressively challenging supervision.
\item We perform detailed quantitative and qualitative evaluations by question type, revealing model-specific strengths, weaknesses, and the effects and tradeoffs of fine-tuning at different scales. The code for implementing our pipeline are publicly available at: \href{https://github.com/eeyore22/urban_scope}{https://github.com/eeyore22/urban\_scope}.
\end{itemize}

\begin{figure*}[h]
    \centering
    \includegraphics[width=1.0\linewidth]{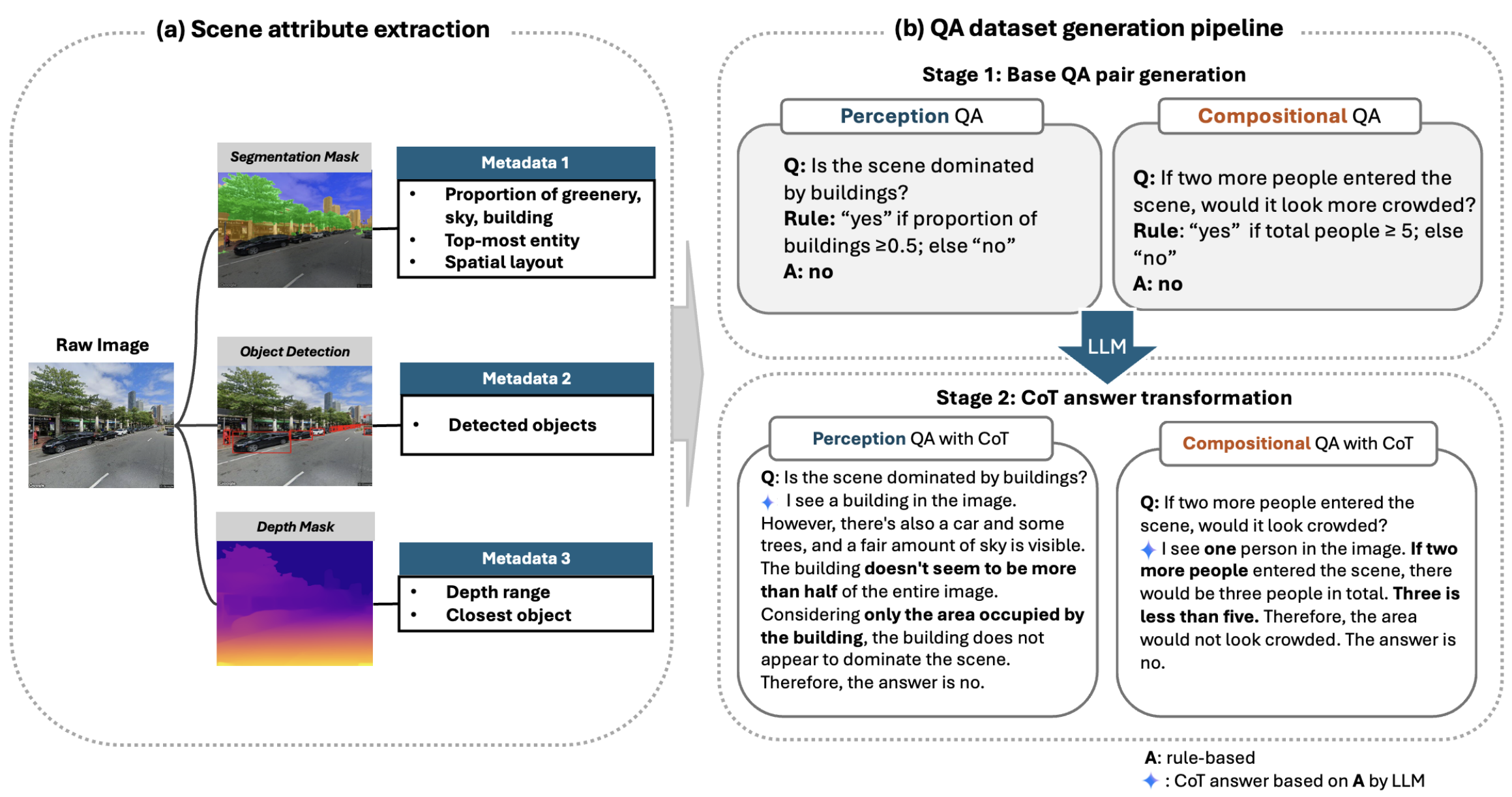}
    \caption{Overview of our pipeline. \textbf{(a)} We take raw street-view images as inputs and extract three key predictions: segmentation masks, object detection results, and monocular depth masks. Each output contributes to the assembled metadata describing the street scene, including view factor proportions (greenery, sky, building), object counts, spatial layout distribution, depth complexity, and closest object information. \textbf{(b)} The QA dataset generation module uses this metadata to generate diverse base QA pairs through perception-guided and compositional-guided components. These metadata-grounded answers provide the foundation for the Chain-of-Thought (CoT) QA transformation stage, which enriches each QA pair with step-by-step natural language reasoning using an LLM.}
    \label{fig:overall_pipeline}
\end{figure*}

  \section{Related Work}

\textbf{Enhancing reasoning in vision–language models.}
A growing body of work has focused on improving the reasoning capabilities of vision–language models. Gokhale et al. (2020) introduced the VQA-LOL benchmark to specifically evaluate logical reasoning in VQA, demonstrating that models frequently fail on tasks involving negation and complex logic~\cite{Gokhale2020}. To address similar challenges, Niu et al. (2021) proposed a counterfactual VQA framework that mitigates language biases, showing that many VQA models rely on spurious correlations rather than true scene understanding~\cite{Niu2021}. Zhang et al. (2025) further emphasized that even modern multimodal models still misinterpret negated or hypothetical questions in urban driving scenarios, underscoring persistent reasoning gaps~\cite{Zhang2025}.

Synthetic and programmatically generated benchmarks have further advanced reasoning evaluation. Johnson et al. (2017) introduced the CLEVR benchmark, which used synthetic 3D scenes to systematically test compositional and logical reasoning in VQA~\cite{Johnson2017}. Hudson and Manning (2019) developed GQA, a large-scale, programmatically generated VQA dataset that specifically targets multi-step reasoning on real images~\cite{Hudson2019}. Chen et al. (2024) introduced SpatialVLM, which augmented VQA training with depth-aware, spatially grounded questions to improve quantitative spatial reasoning~\cite{Chen2024}. Wang et al. (2025) developed OmniDrive, which synthesized counterfactual driving scenarios (e.g., “If I decide to accelerate and make a left turn, what could be the consequences?”) to generate decision-oriented QA pairs tailored for driving applications~\cite{Wang2025}.

\vspace{0.3em}
\noindent
\textbf{Urban scene understanding and computer vision foundations.}
Meanwhile, the computer vision community has developed a wide range of tools and foundational datasets for urban scene understanding. Cityscapes~\cite{Cordts2016} and Mapillary Vistas~\cite{Neuhold2017} remain among the most widely used resources in this domain. Cityscapes provides finely annotated urban street scenes and significantly advanced semantic segmentation in complex driving environments. Mapillary Vistas extends this foundation by offering diverse street-level imagery across a broader range of cities, countries, and viewpoints, enhancing generalization beyond uniform city structures.

Ranftl et al. (2020) proposed MiDaS, a cross-dataset monocular depth estimation framework that improved depth generalization across various visual domains, including urban scenes~\cite{Ranftl2020}. Caesar et al. (2020) introduced nuScenes, which provided multimodal sensory inputs—such as lidar, radar, and multi-camera setups—for urban driving tasks~\cite{Caesar2020}, enabling detailed spatial and temporal scene understanding. These perception pipelines now offer high-quality scene metadata that can support reasoning-centric evaluations in vision–language tasks.

  \vspace{-8pt} 
\section{Methods}
\label{sec:methods}


\subsection{Street-view Image Collection}
We collected 50,000 street view images at a resolution of 512×512 from five global cities—Boston, New York City, Tokyo, Seoul, and Singapore—using the Google Street View API~\cite{Anguelov2010}. Sampling locations were systematically selected via OpenStreetMap (OSM) road network coordinates to ensure coverage of publicly accessible streets. At each location, we captured four headings ($0^\circ$, $90^\circ$, $180^\circ$, and $270^\circ$) to provide a full panoramic view of the urban context. The dataset spans diverse urban typologies, including downtown cores, residential streets, waterfronts, green corridors, and mixed-use areas, enabling broad generalization across spatial environments.

\subsection{Scene Attribute Extraction}
\label{sec:metadata}
For each street-view image $I$, we extract scene attributes using pretrained models: semantic segmentation, object detection, and monocular depth estimation. From these, we assemble structured metadata of each image such as greenery proportion, object counts, and depth range. This metadata serves as pseudo-ground truth, offering scalable, interpretable supervision for spatial reasoning tasks.

\vspace{0.3em}
\noindent 
\textbf{Semantic Segmentation.} We use SegFormer \cite{Xie2021} pretrained on Cityscapes dataset ~\cite{Cordts2016} to obtain pixel-wise class labels and calculate the proportions of greenery, sky, and buildings—often quantified as Green View Index (GVI), Sky View Factor (SVF), and Building View Factor (BVF) \cite{Gong2018, Nagata2020, Li2015}. These metrics help characterize visual structure in urban scenes, enabling tasks like detecting dominant vertical arrangements and left–right spatial bias.

\vspace{0.3em}
\noindent
\textbf{Object Detection.} DETR ResNet-50 \cite{Carion2020} is used to detect key urban scene elements (e.g., pedestrians, cars, bicycles). We extract both counts and bounding box locations, enabling questions about object presence, quantity, and co-occurrence.

\vspace{0.3em}
\noindent
\textbf{Monocular Depth Estimation.} Using MiDaS \cite{Ranftl2020}, we generate per-pixel depth maps and derive scene-level statistics such as depth range, variance, and object-wise depth averages. These inform reasoning about spatial complexity and object proximity.

\vspace{0.3em}

\noindent
\textbf{Metadata Assembly.} The predictions are consolidated into a unified, structured metadata record for each image. This record directly drives base QA pair generation, ensuring that every question is traceable to specific visual evidence in the scene.

Following prior benchmarks such as SpatialVLM \cite{Chen2024} and OmniDrive \cite{Wang2025}, we adopt a synthetic supervision strategy based on pretrained model outputs rather than human annotations. This approach carries the risk of occasional errors such as reduced pixel-level precision, but offers a practical tradeoff: enabling large-scale, consistent QA generation with broad task coverage and reproducibility, thereby supporting more transparent and controlled evaluations of spatial reasoning in vision–language models.

\begin{table}[h]
    \centering
    \caption{Example of extracted metadata for a single street-view image.}
    \begin{tabular}{ll}
        \toprule
        \textbf{Field} & \textbf{Value} \\
        \midrule
        Greenery Proportion & 0.35 \\
        Sky Proportion & 0.15 \\
        Building Proportion & 0.40 \\
        Objects & {person: 2, car: 5, building: 2} \\
        Depth Range & 41.5 \\
        Closest Object & person \\
        Layout & buildings: left, cars: right \\
        Top Entity & building \\
        \bottomrule
    \end{tabular}
    \label{tab:metadata_example}
\end{table}

\subsection{QA Generation Pipeline}
We generate a QA dataset in two main phases: (1) creation of \emph{base QA pairs} with short factual answers, and (2) transformation of the answers into \emph{Chain-of-Thought (CoT) variants} that verbalize the underlying reasoning process. 
Base QA generation covers two broad perspectives: a \textbf{perceptual} perspective, which can be answered directly from a single scene attribute, and a \textbf{compositional} perspective, which requires combining multiple attributes through intermediate logic. 
Appendix 6.1 details the number of QA pairs per question type, with templates and type-specific rules in Appendices 6.2–6.3.

\subsubsection{Base QA Generation}
\noindent
The first phase produces base QA pairs with short, factual answers. 
All questions are grounded in structured metadata of segmentation, object detection, and depth estimation predictions, and fall into two broad categories:

\vspace{-8pt}
\paragraph{Perceptual QA.}
These questions can be answered directly from a single scene attribute, using deterministic rules and interpretable thresholds. 
Answers are typically numeric or binary, reflecting raw scene measurements.

\vspace{3pt}
\begin{itemize}
\item \textbf{Proportions:} Scalar or binary assessments of the pixel-wise proportions of greenery, sky, and buildings.
\item \textbf{Depth:} Identification of the closest object using depth maps.
\item \textbf{Layout:} Inference of vertical composition and left/right dominance.
\item \textbf{Objects:} Counts, presence, and co-occurrence of urban scene elements.
\end{itemize}

\noindent
Perceptual thresholds are drawn from prior studies linking feature proportions (e.g., $ \text{GVI} > 30\% $) to visual dominance and aesthetic appraisal~\cite{Aoki1991, Bolte2024, Zhu2021, Miao2020, Svensson2004}. These perception-level outputs serve as structured evidence for the reasoning and CoT stages, while also enabling standalone evaluation of intuitive visual patterns such as object prominence and spatial asymmetry.


\vspace{-10pt}
\paragraph{Compositional QA.}
These questions require integrating multiple perceptual facts and applying intermediate logic to produce a higher-order answer. 
The output remains short, such as yes/no, integer, or a single word, but the derivation follows a fixed, question type–specific reasoning rule.

\vspace{5pt}
\begin{itemize}
\item \textbf{Negation:} Tests the ability to process exclusions or counter-statements based on perceptual evidence.
\item \textbf{Counterfactuals:} Hypothetical scenarios constructed from plausible alternatives to the observed scene attribute.
\item \textbf{Multi-hop:} Multi-step comparisons and chained logic that traverse multiple perceptual attributes.
\end{itemize}

\noindent
Negation and counterfactuals pose well-known challenges to both humans and AI models~\cite{Dudschig2021, Kulakova2016, Kassner2019}. Prior work has shown that LLMs often misinterpret these forms~\cite{Li2024, Vrabcova2025}, underscoring the need for rigorous evaluation of compositional reasoning grounded in explicit perceptual inputs.


\subsubsection{Chain-of-Thought QA Transformation}
\label{sec:cot_module}
In the second phase, each base QA pair is transformed into a \emph{CoT variant} by replacing its short factual answer with a step-by-step natural language rationale that reconstructs the original reasoning process. “Thinking-aloud” reasoning has been shown to enhance performance and interpretability in tasks such as arithmetic, commonsense inference, and multi-hop reasoning~\cite{Wei2022, Kojima2022, Wang2023}, and has recently been adapted to vision–language settings~\cite{Zhang2024CoT, Shao2024, Chen2024c}.
In our framework, CoT is not an additional question type but an \emph{answer expansion layer} applied to both perceptual and compositional QA.

For each QA pair, Gemini 1.5-Flash~\cite{gemini2024} is prompted with the question, the scene's metadata (Table~\ref{tab:metadata_example}), and the corresponding reasoning rule from a predefined question type–to–reasoning rule mapping table (shown in Appendix 6.4).
The model is instructed to treat the metadata as visual evidence, follow the specified rule, and produce a step-by-step natural language rationale before stating the final answer, ensuring that the explanation remains fully grounded in the scene attributes extracted from Section~\ref{sec:metadata}. Figure~\ref{fig:cot_example} shows an example QA and detailed prompt templates are provided in Appendix 6.4.

\subsubsection{Human Validation of Synthetic Supervision}
\label{sec:manual-verification}

Verifying the quality of generated supervision is essential in synthetic benchmarks, as it impacts the reliability and interpretability of downstream evaluations~\cite{singh2024, maheshwari2024, tang2023}.  
Because our QA pipeline uses pretrained models both to extract structured metadata and to produce CoT reasoning traces, we conducted a human evaluation of 500 randomly sampled QA pairs spanning all question types to check for (1) metadata accuracy and (2) CoT answer consistency and plausibility. Judgments were binary, focusing on perceptual plausibility rather than exact numeric precision (e.g., confirming that ``3 cars'' corresponds to roughly three visible cars, or that a greenery proportion of 0.47 appears visually reasonable).  

Key results indicate that predictions from segmentation and depth estimation exhibited high accuracy (95\% and 94\%, respectively), whereas object detection showed comparatively more errors (88\%), primarily due to over-counting objects not visible to human annotators. CoT reasoning was largely consistent with predefined rules (98\% for consistency); however, in certain cases, the direct transformation of rule-based logic did not yield a fully plausible description of the scene’s complexity (90\% for plausibility). Appendix 6.5 provides the full results with representative examples of both correct and incorrect cases.

\begin{figure}[h]
    \centering
    \begin{tcolorbox}[colback=gray!5!white, colframe=gray!75!black, title=Example QA]
        \textbf{Image ID:} image\_12452532923\_heading90.jpg

        \vspace{4pt}
        \textbf{Metadata:} \\
        \texttt{"objects": \{"car":7, "person":3\}}

        \vspace{4pt}
        \textbf{Question:} Are there more people than cars in the image?

        \vspace{4pt}
        \textbf{Type:} multi-hop

        \vspace{4pt}
        \textbf{Rule:} Compare the number of people and cars. If the number of people is greater, the
        answer is “Yes”. Otherwise, “No”.
    
        \vspace{4pt}
        \textbf{Base Answer:} No

        \vspace{4pt}
        \textbf{CoT Answer:} \\
        I see several cars and several people in the image.  Let's count them. I can see at least seven cars. I can also see at least three people.  Seven cars are more than three people. Therefore, there are more cars than people. Answer: No.

    \end{tcolorbox}
    \caption{Example of a synthesized QA pair. The CoT answer faithfully verbalizes the rule-based derivation process using metadata.}
    \label{fig:cot_example}
\end{figure}

\begin{figure*}[h]
    \centering
    \includegraphics[width=1.0\linewidth]{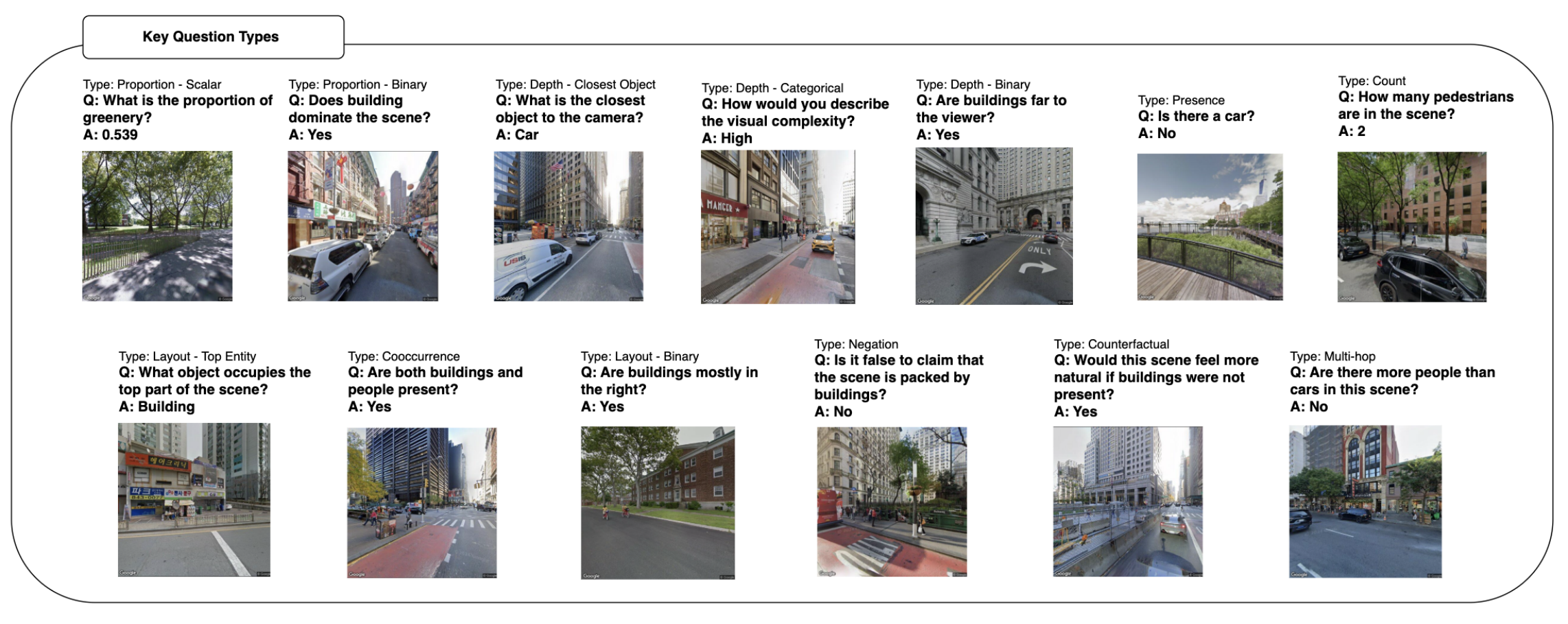}
    \caption{Example question types from the QA generation pipeline. Each example illustrates a representative question, the corresponding street-view image, and the provided answer. For brevity, the full CoT answer is shortened to the final answer.}
    \label{fig:qa_generation_full}
\end{figure*}

\vspace{5pt}
\noindent

  \vspace{-10pt}
\section{Evaluation}
\subsection{Experimental Settings}

We evaluate each question in both zero-shot and fine-tuned settings across BLIP-2 (Flan-T5-xl), InstructBLIP (Flan-T5-xl), and LLaVA-1.5-7B to isolate the effect of task-specific adaptation.

\vspace{0.5em}
\begin{itemize}
    \item \textbf{Zero-shot:} Evaluation on the synthetic VQA dataset without any fine-tuning, measuring each model’s inherent ability to perform spatial reasoning in urban scenes.
    \item \textbf{Fine-tuned:} Evaluation after fine-tuning on the synthetic VQA dataset, measuring how each model’s spatial reasoning performance changes with targeted supervision.
\end{itemize}

\vspace{3pt}
\noindent
All models are fine-tuned using a batch size of 32, for 40 epochs, with a learning rate of 1e-4 and AdamW optimizer. The dataset is split into training, validation, and test sets with a 7:2:1 ratio, and we apply the same splits and hyperparameters across all models for fair comparison. Details regarding the metrics, answer parsing logic, and prompt constraints are in the Appendix 6.6.

\begin{table*}[ht]
  \centering
  \small
  \setlength{\tabcolsep}{4pt}
  \renewcommand{\arraystretch}{1.15}
  \resizebox{\textwidth}{!}{%
  \begin{tabular}{ll
                  c c            
                  c c c          
                  c c            
                  c              
                  c c            
                  c c c}         
  \toprule
  \multirow{3}{*}{\textbf{Model}} & \multirow{3}{*}{\textbf{Setup}}
    & \multicolumn{2}{c}{\textbf{Proportions}}
    & \multicolumn{3}{c}{\textbf{Depth}}
    & \multicolumn{2}{c}{\textbf{Layout}}
    & \multicolumn{3}{c}{\textbf{Object}}
    & \multicolumn{3}{c}{\textbf{Compositional}} \\
  \cmidrule(lr){3-4}\cmidrule(lr){5-7}\cmidrule(lr){8-9}\cmidrule(lr){10-12}\cmidrule(lr){13-15}
    &
    & \multicolumn{1}{c}{Binary} & \multicolumn{1}{c}{Scalar}
    & \multicolumn{1}{c}{Categorical}
    & \multicolumn{1}{c}{Closest Obj.}
    & \multicolumn{1}{c}{Binary}
    & \multicolumn{1}{c}{Binary}
    & \multicolumn{1}{c}{Top entity}
    & \multicolumn{1}{c}{Count}
    & \multicolumn{1}{c}{Presence}
    & \multicolumn{1}{c}{Co-occ.}
    & \multicolumn{1}{c}{Negation}
    & \multicolumn{1}{c}{Counterf.}
    & \multicolumn{1}{c}{Multihop} \\
  \cmidrule(lr){3-3}\cmidrule(lr){4-4}
  \cmidrule(lr){5-5}\cmidrule(lr){6-6}\cmidrule(lr){7-7}
  \cmidrule(lr){8-8}\cmidrule(lr){9-9}
  \cmidrule(lr){10-10}\cmidrule(lr){11-11}\cmidrule(lr){12-12}
  \cmidrule(lr){13-13}\cmidrule(lr){14-14}\cmidrule(lr){15-15}
    &
    & F1↑ & MAE↓
    & F1↑ & F1↑ & F1↑
    & F1↑ & F1↑
    & MAE↓
    & F1↑ & F1↑
    & F1↑ & F1↑ & F1↑ \\
  \midrule
  \multirow{2}{*}{LLaVA-1.5}
    & Zero-shot      & \textbf{0.59} & \textbf{0.18} & \textbf{0.62} & \textbf{0.24} & 0.69 & \textbf{0.61} & 0.09 & \textbf{2.36} & 0.95 & \textbf{0.99} & 0.31 & \textbf{0.35} & 0.36 \\
    & Fine-tuned   & 0.44 & 0.22 & 0.50 & 0.12 & \textbf{0.80} & 0.58 & \textbf{0.15} & 3.03 & \textbf{0.97} & 0.90 & \textbf{0.72} & 0.33 & \textbf{0.40} \\
  \midrule
  \multirow{2}{*}{InstructBLIP}
    & Zero-shot      & \textbf{0.47} & \textbf{0.21} & 0.54 & \textbf{0.22} & \textbf{0.99} & \textbf{0.58} & \textbf{0.03} & 4.33 & 0.96 & \textbf{0.99} & \textbf{0.53} & 0.37 & \textbf{0.80} \\
    & Fine-tuned  & 0.11 & 0.27 & \textbf{0.62} & 0.10 & 0.88 & 0.45 & 0.02 & \textbf{4.05} & \textbf{1.00} & 0.95 & 0.40 & \textbf{0.48} & 0.72 \\
  \midrule
  \multirow{2}{*}{BLIP-2}
    & Zero-shot      & 0.33 & 0.21 & 0.11 & 0.11 & \textbf{0.99} & \textbf{0.58} & 0.06 & 4.10 & 0.97 & 0.91 & 0.52 & 0.55 & 0.67 \\
    & Fine-tuned  & \textbf{0.89} & \textbf{0.11} & \textbf{0.76} & \textbf{0.67} & 0.89 & 0.57 & \textbf{0.87} & \textbf{1.70} & \textbf{0.98} & \textbf{1.00} & \textbf{0.91} & \textbf{0.90} & \textbf{0.81} \\
  \bottomrule
  \end{tabular}}
  \caption{Performance of vision–language models on perceptual QA and compositional QA tasks. Results are reported for zero-shot and fine-tuned settings. Bold indicates the better performance within each model and metric.}
  \label{tab:final_tab}
\end{table*}

\begin{table*}[ht]
  \centering
  \small
  \setlength{\tabcolsep}{4pt}
  \renewcommand{\arraystretch}{1.15}
  \resizebox{\textwidth}{!}{%
  \begin{tabular}{ll
                 c c
                 c c c
                 c c
                 c c c
                 c c c}
  \toprule
  \multirow{2}{*}{\textbf{Model}} & \multirow{2}{*}{\textbf{Setup}}
    & \multicolumn{2}{c}{\textbf{Proportions} ($\% \Delta$)}
    & \multicolumn{3}{c}{\textbf{Depth} ($\% \Delta$)}
    & \multicolumn{2}{c}{\textbf{Layout} ($\% \Delta$)}
    & \multicolumn{3}{c}{\textbf{Object} ($\% \Delta$)}
    & \multicolumn{3}{c}{\textbf{Compositional} ($\% \Delta$)} \\
  \cmidrule(lr){3-4}\cmidrule(lr){5-7}\cmidrule(lr){8-9}\cmidrule(lr){10-12}\cmidrule(lr){13-15}
    & & Binary & Scalar & Categorical & Closest Obj. & Binary & Binary & Top Entity & Count & Presence & Co-occ. & Negation & Counterf. & Multihop \\
  \midrule
  LLaVA-1.5 & $\% \Delta$ & –25.4 & –22.2 & –19.4 & –50.0 & +15.9 & –4.9 & +66.7 & –28.4 & +2.1 & –9.1 & +132.3 & –5.7 & +11.1 \\
  InstructBLIP & $\% \Delta$ & –76.6 & –28.6 & +14.8 & –54.5 & –11.1 & –22.4 & –33.3 & +6.5 & +4.2 & –4.0 & –24.5 & +29.7 & –10.0 \\
  BLIP-2 & $\% \Delta$ & +169.7 & +47.6 & +590.9 & +509.1 & –10.1 & –1.7 & +1350.0 & +58.5 & +1.0 & +9.9 & +75.0 & +63.6 & +20.9 \\
  \bottomrule
  \end{tabular}}
  \caption{Percentage change from zero-shot to fine-tuned for each model and task type. Positive values indicate improvement (higher F1 or lower MAE), while negative values indicate performance degradation.}
  \label{tab:percentage_change}
\end{table*}

\subsection{Quantitative Evaluation}
\subsubsection{Heterogeneous Performance Trends} 
When we compare zero‐shot performance to CoT fine‐tuning across all task types, three distinct patterns emerge. \textbf{Significant improvement} group includes tasks such as \emph{counterfactual reasoning}, \emph{negation reasoning}, and \emph{depth‐categorical} questions. For example, as shown in Table~\ref{tab:percentage_change}, BLIP-2 shows a remarkable 509\% gain in \emph{depth-closest object} and a 591\% improvement in \emph{depth-categorical} questions. \emph{Negation} and \emph{counterfactual} reasoning tasks also see substantial improvements across models, with BLIP-2 achieving a 75\% increase in \emph{negation} and a 64\% increase in \emph{counterfactual} reasoning. These gains suggest that a few thousand in-domain examples can rapidly equip models to address reasoning gaps in urban scenes. \textbf{Marginal improvement} group—most notably \emph{object presence} and \emph{multihop reasoning}—shows more modest gains, with BLIP-2 improving by around 1\% in \emph{object presence} and 21\% in \emph{multihop} reasoning. Similarly, LLaVA-1.5 demonstrates a slight 11\% improvement in \emph{multihop} tasks and minimal gains in \emph{object presence}. This suggests that fine-tuning aids logical chaining but may not fully overcome the inherent challenges in these complex, compositional tasks. 

\textbf{Performance degradation} group includes tasks such as \emph{object co-occurrence}, \emph{layout binary}, and \emph{proportion binary}, where models frequently show drops after fine-tuning. For instance, LLaVA-1.5 and InstructBLIP experience a 9--28\% decrease in \emph{object co-occurrence} and \emph{proportion binary}, suggesting that adaptation to in-domain dataset can sometimes erode zero-shot strengths in simpler perceptual tasks. One plausible explanation is \textit{catastrophic forgetting}, where fine-tuning on a dataset dominated by compositional and reasoning-heavy samples shifts the model’s feature representations away from low-level perceptual cues. A second factor could be \textit{data distribution mismatch}, as our fine-tuning set contains relatively fewer straightforward perception cases. Our fine-tuning also emphasizes rule-based reasoning traces, which could draw attention toward abstract scene logic at the expense of rapid, single-step perceptual judgments. These findings suggest pairing domain-specific adaptation with strategies such as rehearsal data, multi-task balancing, or selective freezing to preserve perceptual competence while boosting reasoning performance.

\subsubsection{Model Efficiency and Practical Trade-offs}
As shown in Figure~\ref{fig:params_efficiency}, BLIP-2 offers the most parameter-efficient gains, achieving substantial perception and reasoning improvements despite being the smallest model in this comparison. In contrast, LLaVA-1.5 specializes in reasoning tasks, showing strong reasoning gains with minimal parameter overhead but a noticeable decline in perception tasks, such as a 25\% drop in \emph{proportion binary} and a 28\% increase in \emph{proportion scalar} MAE. Notably, LLaVA-1.5’s robust zero-shot performance makes it an appealing option for deployment in settings with minimal or no task-specific supervision. InstructBLIP, despite its larger model size, demonstrates limited parameter efficiency and suffers from perceptual performance degradation across several tasks, including a 77\% decrease in proportion binary and a 29\% increase in scalar MAE, suggesting that more targeted fine-tuning strategies may be required to fully leverage its potential. Taken together, these results emphasize the need for careful model selection based on the target domain and computational constraints. BLIP-2 emerges as an effective, lightweight model for domain-specific fine-tuning pipelines like ours, providing strong perception and reasoning gains with low computational cost, while LLaVA-1.5 remains a valuable option in reasoning-focused applications, especially when fine-tuning budgets are limited and zero-shot robustness is critical.

\begin{figure}[h]
    \centering
    \includegraphics[width=1.0\linewidth]{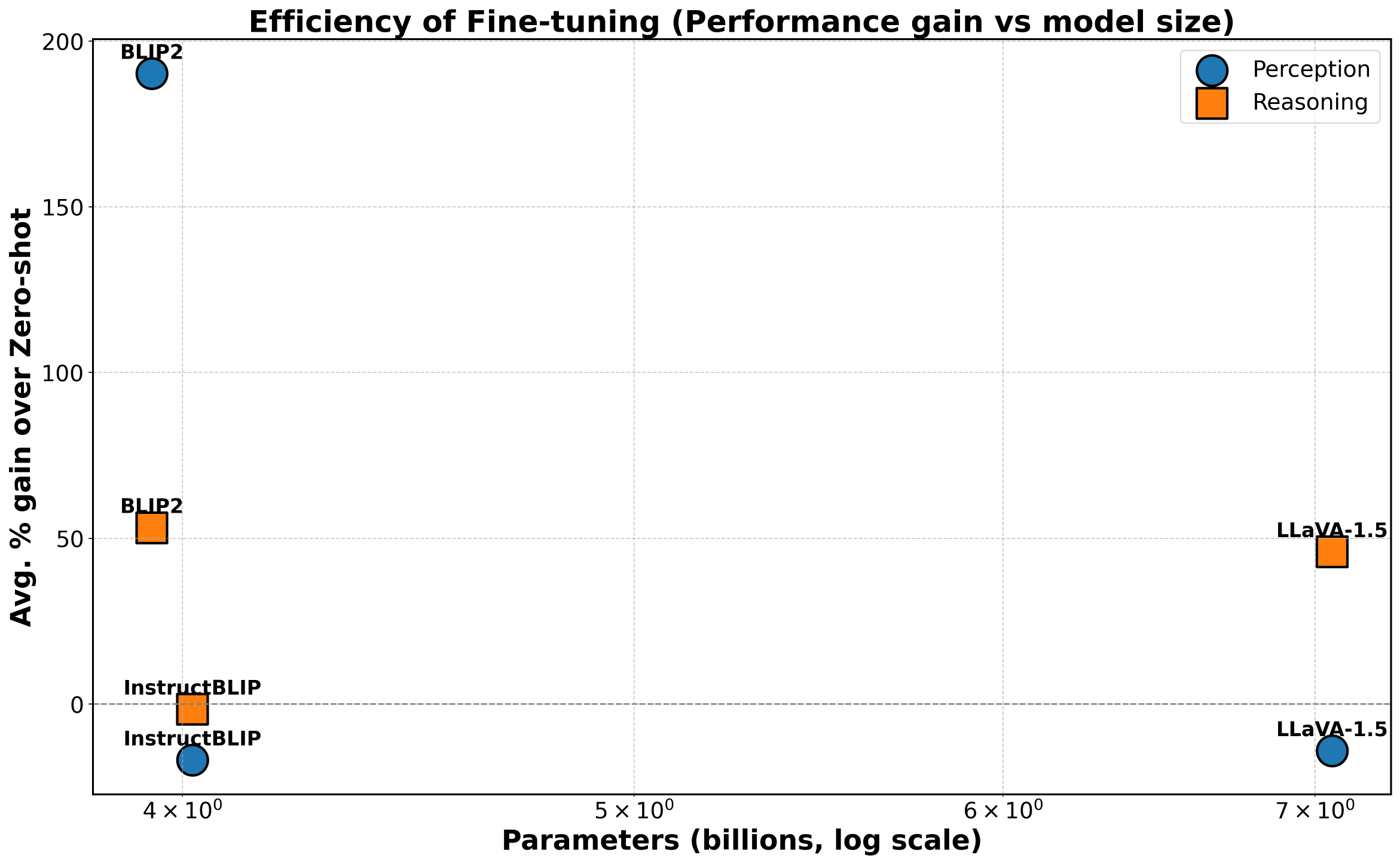}
    \caption{Parameter efficiency of vision–language models based on fine-tuning gains. The x-axis shows the model size in billions of parameters (log scale): LLaVA-1.5 (7.06B), InstructBLIP (4.02B), and BLIP-2 (3.94B). The y-axis shows the average percentage performance gain after fine-tuning compared to zero-shot performance.}
    \label{fig:params_efficiency}
\end{figure}

\subsection{Qualitative Evaluation}
Beyond numerical performance, our qualitative evaluation highlights that different models exhibit distinct strengths across question types. Notably, BLIP-2 consistently demonstrates versatility and robust reasoning capabilities across a wide range of tasks. For example, as illustrated in Figure~\ref{fig:qualitative_evaluation}, in a representative negation-type question, fine-tuning enhances reasoning quality for both InstructBLIP and BLIP-2 compared to their zero-shot counterparts. However, while InstructBLIP tends to remain at the level of question repetition—stating that ``the statement says that neither buildings nor sky dominates this scene''—BLIP-2 engages in more explicit logical reasoning, accurately resolving the double-negative structure of the question by deducing: ``while neither the sky nor the building takes up the majority of the image, they are both present and visible. Therefore, it’s false to say that neither dominates the scene.'' 

While LLaVA-1.5 exhibits comparatively lower average performance, as indicated by its F1 scores in Table~\ref{tab:final_tab}, it nonetheless produces occasional high-quality answers. We report one example of a depth task in Figure~\ref{fig:qualitative_evaluation}. The model consistently provides detailed and contextually appropriate assessments of scene depth, even when other models struggle. Additionally, InstructBLIP shows notable proficiency in counterfactual reasoning. Fine-tuning substantially improves its ability to perform multi-step hypothetical reasoning beyond surface-level descriptions, as shown in the counterfactual example in the same figure.

Collectively, these qualitative observations complement the quantitative results, indicating that while fine-tuning with a domain-specific synthetic QA dataset generally improves model performance, each model retains distinct strengths more apparent through qualitative analysis.

\begin{figure*}[ht]
\centering
\begin{adjustbox}{width=0.98\textwidth}
\renewcommand{\arraystretch}{1.35}
\begin{tabular}{%
  >{\centering\arraybackslash}m{0.17\textwidth}  
  >{\raggedright\arraybackslash}p{0.17\textwidth}
  >{\centering\arraybackslash}p{0.09\textwidth}  
  >{\centering\arraybackslash}p{0.09\textwidth}  
  >{\raggedright\arraybackslash}p{0.48\textwidth}}
\toprule
\textbf{Image} & \textbf{Question} & \textbf{Model} & \textbf{Setting} & \textbf{Answer}\\
\midrule

\multirow{10}{=}{\includegraphics[width=0.95\linewidth]{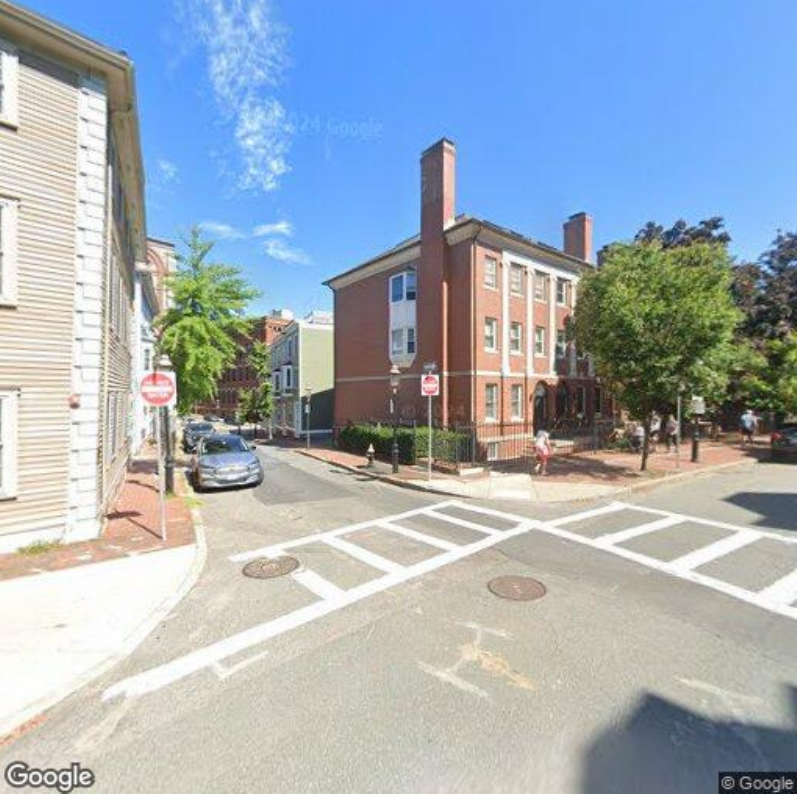}}
& \multirow{10}{=}{\textbf{Q:} If two more people entered the scene, would it look crowded? Respond "yes" or "no".}
& \multirow{2}{*}{InstructBLIP} & Zero-shot & no \\
& &                            & Fine-tuned & I see four people in the image. If two more people were to enter the scene, there would be six people in total. Six is greater than or equal to five. Therefore, the area would look crowded. answer: yes \\ \cmidrule(lr){3-5}

& & \multirow{2}{*}{BLIP-2} & Zero-shot & if two more people entered the scene would it look crowded \\
& &                        & Fine-tuned & i see one person in the image. if two more people entered the scene, there would be a total of three people. three is less than five. therefore, the scene would not look crowded. answer: no \\ \cmidrule(lr){3-5}

& & \multirow{2}{*}{LLaVA-1.5} & Zero-shot & No \\
& &                            & Fine-tuned & yes \\ \cmidrule(lr){3-5}

& & \multicolumn{2}{r}{\textbf{Ground-truth}} & \textbf{Yes}\\
\midrule

\multirow{10}{=}{\includegraphics[width=0.95\linewidth]{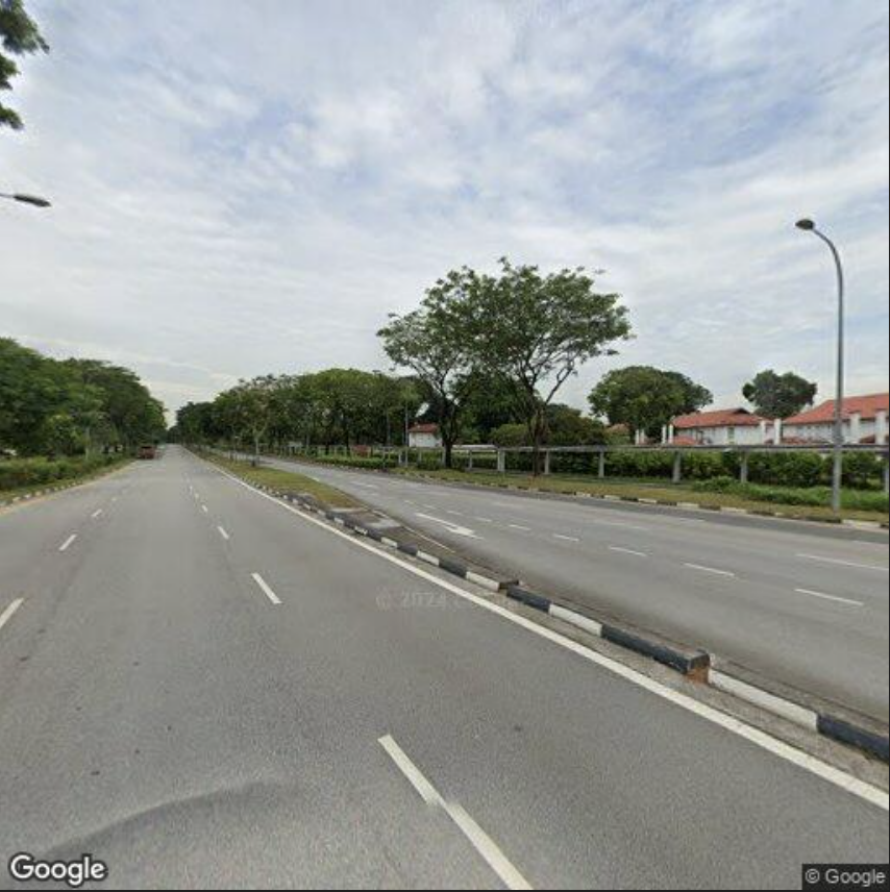}}
& \multirow{10}{=}{\textbf{Q:} Is it false to say that neither buildings nor sky dominate this scene? Respond in "yes" or "no".}
& \multirow{2}{*}{InstructBLIP} & Zero-shot & No \\
& &                            & Fine-tuned & the image shows a scene with a building and sky dominating the scene. the statement says that neither buildings nor sky dominates this scene. answer: no \\ \cmidrule(lr){3-5}

& & \multirow{2}{*}{BLIP-2} & Zero-shot & No  \\
& &                        & Fine-tuned & looking at the image, i see a significant portion of the sky at the top. there's also a building visible on the right side. while neither the sky nor the building takes up the majority of the image, they are both present and visible. therefore, it's false to say that neither dominates the scene. answer: no \\ \cmidrule(lr){3-5}

& & \multirow{2}{*}{LLaVA-1.5} & Zero-shot & No \\
& &                            & Fine-tuned & no no no no no (...omitted repetitive text...) answer: no\\ \cmidrule(lr){3-5}

& & \multicolumn{2}{r}{\textbf{Ground-truth}} & \textbf{No}\\
\midrule

\multirow{10}{=}{\includegraphics[width=0.95\linewidth]{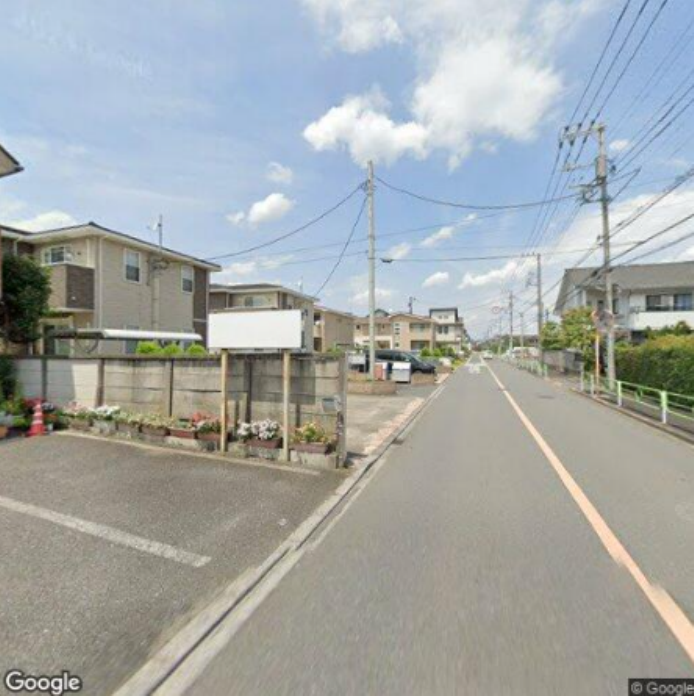}}
& \multirow{10}{=}{\textbf{Q:} Does the scene appear visually complex? Answer in yes or no.}
& \multirow{2}{*}{InstructBLIP} & Zero-shot & Low is the overall depth complexity of the scene, as there are only a few objects in the scene. There is a low amount of detail in the scene, which can be attributed to the low number of objects in the scene. (...)\\
& &                            & Fine-tuned & the looking at the overall depth complexity of this scene. label the overall depth complexity of this scene label the overall depth complexity of this scene label the overall depth complexity of this scene (...) answer: (n/a)\\ \cmidrule(lr){3-5}

& & \multirow{2}{*}{BLIP-2} & Zero-shot & low \\
& &                        & Fine-tuned & looking at the scene, i observe a significant variation in the distances of objects from the camera. the depth range appears quite extensive. the depth complexity is labeled 'high'. answer: high\\ \cmidrule(lr){3-5}

& & \multirow{2}{*}{LLaVA-1.5} & Zero-shot & Moderate\\
& &                            & Fine-tuned & since the depth of the scene is dominated by a wide, clear, and unobstructed road with several cars and a few potted plants, there's not much complexity. the depth range is quite large, exceeding 20 units.  therefore, the depth complexity is high. answer: high \\ \cmidrule(lr){3-5}

& & \multicolumn{2}{r}{\textbf{Ground-truth}} & \textbf{High}\\
\midrule
\end{tabular}
\end{adjustbox}
\caption{Qualitative answers from three VLMs—InstructBLIP, BLIP-2, and LLaVA-1.5—under zero-shot and fine-tuned settings for representative question types.}
\label{fig:qualitative_evaluation}
\end{figure*}

\section{Conclusions and Future Work}

In this study, we evaluated the ability of general-purpose vision-language models to understand fine-grained spatial relationships in street-view images. By introducing a structured pipeline for generating diverse, spatially grounded QA tasks, our work establishes a new problem domain in VL research and creates opportunities for advancing domain-specific perception and reasoning.

Comprehensive experiment results show that fine-tuning with our synthetic QA dataset leads to substantial performance gains. Lightweight models like BLIP-2 particularly benefit from this structured supervision, achieving gains in perception and reasoning capabilities with minimal task-specific data. These findings highlight the potential of synthetic QA with CoT supervision as a versatile approach for enhancing spatial understanding in urban scenes and other domain-specific contexts. For larger models like LLaVA-1.5, our study suggests that synthetic QA alone may be insufficient to shift its pretrained distribution. Addressing this will likely require more complex QA structures and advanced instruction tuning methods, presenting a key direction for future research.

\section*{Acknowledgments}
This work was supported by the National Research Foundation of Korea (NRF) grant funded by the Korea government (MSIT) (No.2022M3J6A1063021, No.RS-2025-00517342).

  \clearpage
  {\small
    \bibliographystyle{ieeenat_fullname}
    \bibliography{main}
  }
\fi

\end{document}